\def\year{2020}\relax
\documentclass[letterpaper]{article} 
\usepackage{aaai20}  
\usepackage{times}  
\usepackage{helvet} 
\usepackage{courier}  
\usepackage[hyphens]{url}  
\usepackage{graphicx} 
\usepackage{graphicx}  
\newif\ifdebugdoc\debugdoctrue

\ifdebugdoc

\newcommand{\fyi}[1]{\footnote{\textcolor{blue}{fyi:#1}}}

\newcommand{\add}[1]{\textcolor{red}{#1}}
\newcommand{\del}[1]{\textcolor{blue}{\sout{#1}}}
\newcommand{\outline}[1]{\textbf{\colorbox{yellow}{Outline:}\textcolor{red}{#1.}}}

\newcommand{\jie}[1]{\footnote{\colorbox{yellow}{Jie:} #1.}}

\else

\newcommand{\fyi}[1]{}

\newcommand{\add}[1]{#1}
\newcommand{\del}[1]{}
\newcommand{\jie}[1]{}
\newcommand{\outline}[1]{}
\fi

\newcommand{\nop}[1]{}

\newcommand{\smsuperscript}[1]{\text{\scaleto{\text{(#1)}}{5pt}}}

\usepackage{xcolor}
\usepackage{balance}
\usepackage{url}
\usepackage{enumitem}
\usepackage{amsmath}
\usepackage{booktabs}
\usepackage{multirow}
\usepackage{graphicx}
\usepackage{soul}
\usepackage{amsthm}
\theoremstyle{definition}
\newtheorem{definition}{Definition}[]

\usepackage{makecell}
\usepackage{cases}
\usepackage{scalerel}
\newcommand{\textlangle}{$\langle$}
\newcommand{\textrangle}{$\rangle$}

\newcommand{\citet}[1]{\citeauthor{#1} \shortcite{#1}}
\newcommand{\citep}{\cite}

\title{Easy-to-Hard: Leveraging Simple Questions for Complex Question Generation}
 \author{Jie Zhao,
Xiang Deng,
Huan Sun,\\
The Ohio State University\\
\{zhao.1359, deng.595, sun.397\}@osu.edu}
\begin{document}

\maketitle

\begin{abstract}
This paper makes one of the first efforts toward automatically generating \emph{complex} questions from knowledge graphs. 
Particularly, we study how to leverage existing simple question datasets for this task, under two separate scenarios: using either sub-questions of the target complex questions, or distantly related pseudo sub-questions when the former are unavailable.
First, a competitive base model named \textsf{\small CoG2Q} is designed to map \ul{co}mplex query \ul{g}raphs to natural language \ul{q}uestions.
Afterwards, we propose two extension models, namely \textsf{\small CoGSub2Q} and \textsf{\small CoGSub$^m$2Q}, respectively for the above two scenarios.
The former encodes and copies from a sub-question, while the latter further scores and aggregates multiple pseudo sub-questions.
Experiment results show that the extension models significantly outperform not only base \textsf{\small CoG2Q}, but also its augmented variant using simple questions as additional training examples.
This demonstrates the importance of \emph{instance-level} connections between simple and corresponding complex questions, which may be underexploited by straightforward data augmentation of \textsf{\small CoG2Q} that builds \emph{model-level} connections through learned parameters.
\end{abstract}

\section{Introduction}
\label{sec:intro}
In this work, we study the task of Question Generation (QG) from Knowledge Graph (KG) queries.
This task, per se, reflects the ability of intelligent systems to translate abstract and schema-specific KG representations into universally understandable natural language questions.
Such automatically generated questions can be used to enrich human-machine conversation and facilitate the scenario where machines need to seek information from users \cite{Wang2018LearningTA,Gur2018DialSQLDB}.
They may also be used as additional data to train question answering systems to save the cost of expensive human annotations \cite{serban2016generating,Jain2017CreativityGD,Guo2018QuestionGF}.

\begin{figure}
    \centering
    \includegraphics[width=\linewidth]{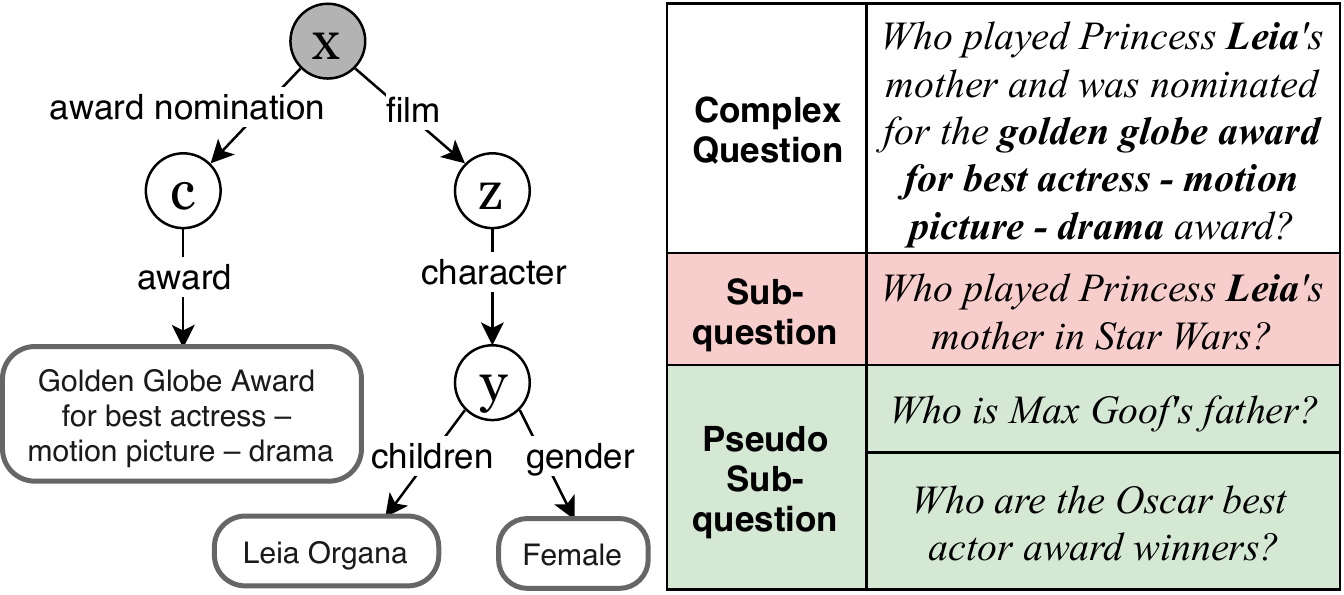}
    \caption{An example of a complex query graph and question pair, with its \textit{sub-question} and two \textit{pseudo sub-questions} (See Definition~\ref{def:subq} and \ref{def:pseudo-subq} in Section~\ref{sec:preliminary}).}
    \label{fig:example}
\end{figure}

For example, the left side of Figure~\ref{fig:example} shows a KG query represented as a graph with $x$ being the question node.
It has the same meaning as the complex question on the right.
As demonstrated here, KG queries are known for the expressiveness of their semantic representations through the combination of different relations between different entities.
However, to the best of our knowledge, existing work on question generation are limited to generating questions from \emph{simple} KG queries each containing only one relation triple, e.g., from (\texttt{\small{Padm\'e Amidala}}, \texttt{\small{children}}, \texttt{\small{Leia Organa}})\nop{(\texttt{\small{Leia Organa}}, \texttt{\small{parent}}, \texttt{\small{Padm\'e Amidala}})} to ``\textit{Who is Queen Amidala's child?}".
Hence, we want to study QG in the more general setting, and design a system that can handle arbitrary, and especially \emph{complex} KG queries with more than one relations.

Complex Question Generation (CQG) is more challenging than simple QG, because the relations not only need to be expressed in natural language correctly, they also have to be fused naturally and coherently. 
It is also difficult to manually design rules to combine simple questions into complex ones, because KG comes with different types and granularity of compositions.
For example in Figure~\ref{fig:example}, the sub-tree under node $y$ contains two relations showing that $y$ has \texttt{\small{children}} and $y$'s \texttt{\small{gender}} is female.
These two relations of $y$ are often expressed together in natural language with a single word ``\textit{mother}".
Moreover, the left sub-tree under $x$ contains two relations \texttt{\small{award\_nomination}} and \texttt{\small{award}} connected by node $c$, which has the specifically designed Compound Value Type (CVT) that allows KG to represent a multi-argument relation\nop{which seem duplicated but actually are specifically designed to represent multi-argument relations in KG} -- e.g. an award nomination event that can be related to a film, a person, a reward, a time, etc.

We first design a base CQG model named \textsf{\small CoG2Q}\nop{\footnote{which stands for \ul{Co}mplex query \ul{G}raph to \ul{Q}uestion.}}, which is an encoder-decoder neural network
following the methodology of state-of-the-art simple QG work \cite{serban2016generating,elsahar2018zero}.
It employes a tree-LSTM \cite{tai2015improved} encoder that can handle arbitrary query graphs, and an LSTM \cite{Hochreiter1997LongSM} decoder with attention over individual relations.

Neural methods are often data hungry and unfortunately, for CQG, there is a lack of training instances to cover the exponentially growing number of different relation combinations.
Therefore, it is desirable to effectively leverage existing simple questions to train the model.
In particular, we examine two separate scenarios:
(1) where the \emph{sub-questions} of the to-be-generated complex question are available {for training}, and (2) where sub-questions are rare, but there are easy-to-find but distantly related \emph{pseudo sub-questions} about other entities.

A straightforward strategy to improve the CQG performance with simple questions is data augmentation, i.e., using them as additional training instances.
This strategy \nop{is well suited for the one-size-fits-all}can be directly applied to the \textsf{\small CoG2Q} model, which handles simple and complex question generation indiscriminately.
However, it only establishes \emph{model-level} connections between \emph{all} the simple and \emph{all} the complex questions through jointly fitted model parameters, which may be sub-optimal due to the underexploitation of the individual correspondence between a complex question and its (pseudo) sub-questions.

In contrast, we further proposed two extension models \textsf{\small CoGSub2Q} and \textsf{\small CoGSub$^m$2Q} that exploit \emph{instance-level} connections between simple and complex questions.
\textsf{\small CoGSub2Q} is used in the first scenario, which additionally encodes a sub-question and explicitly copies words from it while generating the target complex question.
{This is analogous to how humans would sometimes write a sentence with complex logic: first have in mind a simple sentence that covers part of the logic and then build upon it to form the complex sentence.}
\textsf{\small CoGSub$^m$2Q} is designed for the second scenario where pseudo sub-questions are used as surrogates for sub-questions.
To compensate for the lack of actual sub-questions, \textsf{\small CoGSub$^m$2Q} simultaneously takes as input multiple pseudo sub-questions, which are jointly scored and aggregated to help generate the final complex question.
{Back to the human writing scenario, this is analogous to that where one may first look up a list of relevant simple expressions{ about a topic} and then pick the most suitable one(s) to use.}
Experimental results on the ComplexWebQuestions \cite{talmor2018web} dataset show that \textsf{\small CoGSub2Q} and \textsf{\small CoGSub$^m$2Q} can more effectively utilize sub-questions and multiple pseudo sub-questions than straightforward data augmentation, and generate better complex questions.\nop{can utilize sub-questions and multiple pseudo sub-questions to generate better complex questions, which is more effective than straightforward data augmentation}

\section{Preliminaries}
\label{sec:preliminary}
\noindent \textbf{Knowledge Graph.} A knowledge graph $\mathcal{K}$ is a large-scale directed graph where nodes are \textit{entities} and edges are \textit{predicates} describing the relationships between entities.
In practice, a KG is usually represented as a set of relation triples $(s, p, o) \in \mathcal{E}\text{$\times$}\mathcal{P}\text{$\times$}\mathcal{E}$, where $\mathcal{E}$ is the set of all entities and $\mathcal{P}$ is the set of all predicates.
In this work we use Freebase \cite{bollacker2008freebase}, which \nop{is one of the most popular KGs and}has millions of nodes and billions of relations, and is widely adopted by previous works for KG based question generation as well as various other tasks.

\noindent \textbf{Query Graph.} A query graph is also a connected and directed graph, but usually much smaller than a KG.
It can be represented by a relation triple set $G\text{$=$}\{r_i\}_{i=1:N}$, $r_i\text{$=$}(s_i,p_i,o_i)$. 
Figure~\ref{fig:example} shows an example, which contains relation triples such as ($x$, \texttt{\small{film}}, $z$), ($z$, \texttt{\small{character}}, $y$), ($y$, \texttt{\small{children}}, \texttt{\small{LeiaOrgana}}), etc. $G$ may contain three types of nodes. The shaded node $x$ is the \textit{question} node. $c, z, y$ are \textit{variable} nodes. The rounded rectangles are \textit{terminal} nodes that are already grounded to KG entities.
Nonterminal nodes (e.g. $x,c,y,z$) can be grounded to KG by executing the query graph. We compare their grounded entities when defining \textit{sub-graphs} below (in Def.\ref{def:subq}).
\nop{In practice, a KG can be stored and managed by triple stores (e.g. Virtuoso) and the query graphs can be easily formatted into query languages such as SPARQL to search the KG.}

\noindent \textbf{Problem Definition.} 
\nop{KG-based question generation is to generate a natural language question from a query graph.} Given a query graph $G$, the task of question generation is to output a natural language question $q$ (i.e., a sequence of word tokens $[y_1 ...  y_M]$), which has the same meaning of $G$.
In this work, we focus on complex question generation (CQG) for $G$ with $N\text{$=$}|G|\text{$>$}1$, while previous works focus on \textit{simple} question generation for $G$ with only one relation triple (i.e., $N\text{=}1$).

In particular, we study how to effectively train a CQG model\nop{leverage existing simpler questions for CQG} under two separate scenarios: (1) with \emph{sub-questions} that cover part of $G$
and (2) with \emph{pseudo sub-questions} that are distantly related by only predicates, and much easier to find than strict sub-questions.
At test time, only pseudo sub-questions are assumed available.
(Pseudo) sub-questions are formally defined below.
\theoremstyle{definition}
\begin{definition}[sub-question]
\label{def:subq}
If $G^{sub}=\{r_j\} \subset G$ denotes a \textit{sub-graph} of $G$, we refer to a natural language question $q^{sub}=[z_1 ... z_M]$ corresponding to $G^{sub}$ as a \emph{sub-question} of the to-be-generated $q$ corresponding to $G$.
\end{definition}
\theoremstyle{definition}
\begin{definition}[pseudo sub-question]
\label{def:pseudo-subq}
If ${G^{sub'}} \not\subset G$, but all predicates of ${G^{sub'}}$ appear in $G$, i.e. $\{p_i|(s_i',p_i,o_i') \in {G^{sub'}}\} \subset \{p_i | (s_i,p_i,o_i) \in G\}$, we refer to ${q^{sub'}}$ corresponding to ${G^{sub'}}$ as a \emph{pseudo sub-question} of $q$.. 
\end{definition}


\nop{Noticing that given a complex graph query, its  sub-questions are not readily available for use, we also investigate how to utilize existing simple question datasets (e.g. SimpleQuestions to be introduced in Section \ref{sec:dataset}) to assist CQG. We refer to this setting as \textit{CQG with pseudo sub-questions}. }
\section{Datasets}
\label{sec:dataset}
In this work, we use three datasets whose query graphs are all based on Freebase, but with different levels of question complexity.

\noindent\textbf{WebQuestionsSP (WebQ)} \cite{yih2016value} and \textbf{ComplexWebQuestions (CompQ)} \cite{talmor2018web}. WebQ contains 4,737 real-life questions originally collected using Google suggest and then manually annotated with their matched query graphs\footnote{Query graph, logical form and SPARQL query  are different formalizations that can be transformed into each other. We use fyzz (\url{https://pypi.org/project/fyzz/}) to parse the SPARQL queries in the raw datasets and filter out the ones that can not be parsed.}.
Although most questions in WebQ are simple, their matched query graphs may contain more than one relation, mainly because of Compound Value Type (CVT) nodes for describing multi-argument relations.
For example, $c$ and $z$ in Figure~\ref{fig:example} are CVT nodes.
Each of them is connected to two edges/predicates that jointly represent one relationship: \textit{person nominated for award} for $c$ and \textit{person played character} for $z$.

CompQ is built from WebQ by adding more relations to the original query graphs. A set of 687 manually annotated templates (covering 462 unique KG predicates) are used to combine an original simple question with new relations into a new complex question that is understandable to humans.
Afterwards, crowdsourcing workers are hired to paraphrase these new yet template-based questions into natural complex questions. In total, CompQ consists of 34,689 complex questions\nop{ and XXX unique graph queries}, which have their sub-questions in WebQ because of the relation between these two datasets.
\nop{
\noindent\textbf{ComplexWebQuestions (CompQ)} \cite{talmor2018web}.
This dataset was built for the task of \textit{complex} question answering. It is expanded from WebQ (will be introduced next) by adding more relations to the original query graphs.
A set of 687 manually annotated templates (covering 462 unique KG predicates) were used to combine the original simple question with the new relations into a new complex question that is understandable to humans.
Afterwards, crowdsourcing workers were hired to paraphrase these new yet template-based questions into natural complex questions, and will receive a bonus if their paraphrase differs enough from the template-based generations. In total, CompQ consists of 34,689 complex questions\nop{ and XXX unique graph queries}.
\noindent\textbf{WebQuestionsSP (WebQ)} \cite{yih2016value}. 
This dataset contains real-life questions collected using Google suggest and manually annotated query-graphs.
Although WebQ contains mostly short and simple question, their matched query graphs may contain more than one relations so the previous QG methods cannot be applied. 
This is mainly because of special Compound Value Type (CVT) nodes for describing multi-argument relations in the underlying KG (i.e., Freebase \cite{bollacker2008freebase}).
For example, an Olympic game event is represented by a CVT node with several relations for its year, host city, participant nations etc. 
WebQ consists of 4,737 questions in total.
\add{Complex questions in CompQ can find sub-questions from CompQ because of the relation between these two datasets.}
}

\noindent\textbf{SimpleQuestions(v2) (SimpQ)} \cite{weston2015towards}. This dataset has been widely studied in previous question generation works \cite{serban2016generating,elsahar2018zero}.
It consists of 108,442 natural language questions written by human annotators, which can be answered by a single relation triple in the KG.
{Because they are separately curated, }SimpQ contains many \textit{pseudo sub-questions} for CompQ but few real sub-questions, as it is much easier to find predicates overlapped between a query graph in SimpQ and that in CompQ, than both predicates and entities.
\begin{table}[]
\centering
\caption{Dataset statistics after query graph grouping. For all three datasets, numbers of data samples reduce substantially and numbers of questions per sample are highly screwed. Only 441 (out of 1,837 in total) predicates in SimpQ overlap with CompQ.}
\resizebox{0.99\linewidth}{!}{
\begin{tabular}{@{}p{1.0cm}>{\raggedleft\arraybackslash}p{1.2cm}|>{\raggedleft\arraybackslash}p{0.8cm}>{\raggedleft\arraybackslash}p{0.8cm}>{\raggedleft\arraybackslash}p{0.8cm}|>{\raggedleft\arraybackslash}p{0.8cm}>{\raggedleft\arraybackslash}p{0.8cm}>{\raggedleft\arraybackslash}p{1.0cm}@{}}
\toprule
\multirow{2}{*}{\textbf{Dataset}} & 
\multirow{2}{*}{\begin{tabular}[c]{@{}c@{}}\textbf{No.}\\\textbf{samples}\end{tabular}} & 
\multicolumn{3}{r|}{\textbf{No. rels. per sample}} & 
\multicolumn{3}{r}{\textbf{No. ques. per sample}}\\
        &       & \textbf{Med} & \textbf{Avg}  & \textbf{Max} & \textbf{Med} & \textbf{Avg} & \textbf{Max} \\ \midrule 
WebQ    & 585   & 2     & 1.86  & 5     & 2     & 7.29  & 185   \\
SimpQ   & 441   & 1     & 1.00  & 1     & 20    & 178   & 3,810 \\
CompQ   & 6,162 & 3     & 3.23  & 7     & 2     & 4.09  & 136   \\
\bottomrule
\end{tabular}
}
\label{tab:data_stats}
\end{table}

\noindent\textbf{Data pre-processing and partition.}
We are going to generate questions for query graphs in CompQ by leveraging the sub-questions in WebQ or the pseudo sub-questions in SimpQ as training examples.
{This allows this work not limited to specific training data as pseudo sub-questions are easily available.}

We use entity placeholders during question generation following previous work \cite{serban2016generating}: the appearances of grounded entity names in the questions are replaced by special placeholder tokens\footnote{We follow \citet{serban2016generating} and use difflib: \url{https://docs.python.org/2/library/difflib.html}}.
The CQG models generate questions with indexed placeholders, which can be replaced by the corresponding entity names in post-processing.
This method does not affect the naturalness or grammaticality of generated questions\nop{and also helps reduce ambiguity by using canonical names instead of informal, abbreviated names}. 
In each dataset above, there are many questions with exactly the same query graph except for the grounded entities\footnote{Treating them as different examples is suitable for the questing answering task (which is what these datasets were originally curated for) as they stand for different entity linking examples.}.
After placeholder replacement, these query graphs become identical.
\textit{To avoid duplication for question generation}, we reform the datasets by grouping same query graphs (after entities are replaced by placeholders)\nop{except for entities} into a single sample, and treat their different questions as multiple ground-truth questions.
This gives us multiple references for each query graph, which can make the evaluation more robust. 
We divide the complex query graph groups associated with different sub-questions from WebQ into train/dev/test splits following roughly a 70/15/15 ratio.
The statistics of our datasets are shown in Table~\ref{tab:data_stats}.

\section{\textsf{CoG2Q}: Base Model for CQG}
\label{T2S}
\begin{figure}
    \includegraphics[width=0.98\linewidth]{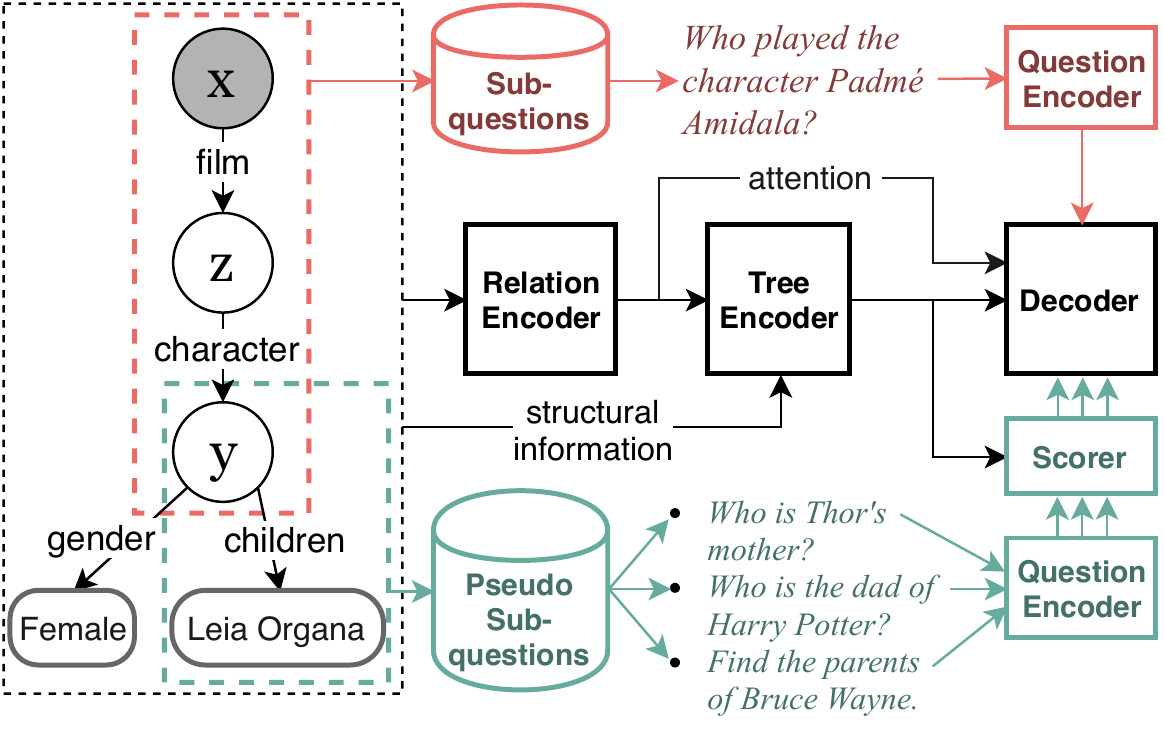}
    \centering
    \caption{Overview of our methodology. Best viewed in color. The middle black components belong to the base model \textsf{\small CoG2Q} (Section~\ref{T2S}).
    The upper red and lower green components are additional ones for \textsf{\small CoGSub2Q} (Section~\ref{sec:incorp-subq}) and \textsf{\small CoGSub$^m$2Q} (Section~\ref{sec:pseudo-subq-cpq}) respectively.
    }
    \label{fig:model}
\end{figure}

We first introduce a strong base model, \textsf{\small CoG2Q}, that takes a query graph as input and outputs a natural language question using the popular encoder-decoder paradigm. 
The encoder in \textsf{\small CoG2Q} is adapted from Tree-LTSM \cite{tai2015improved}  with our unique way to represent relation triples, while the decoder in \textsf{\small CoG2Q} employs an LSTM network \cite{hochreiter1997long} with a novel strategy to generate entity placeholders integrated with an attention mechanism over relation triples.
The black components in Figure~\ref{fig:model} show an overview of \textsf{\small CoG2Q}.

\subsection{Encoder}
\noindent\textbf{Tree-LSTM for encoding a query graph.} Let us assume that each relation triple has been encoded as a vector for now. The tree-LSTM network \cite{tai2015improved} will use them to generate the encoding of the entire query graph and preserve its structural information. We treat the question node as root, and use topological sort to break any circles in the query graph to make it into a tree (duplicating nodes if necessary). Generally, a tree-LSTM aggregates information from leaf nodes to the root, and the hidden states of a node incorporates that of all its child nodes. Specifically in our case:
\begin{equation}
    \label{eq:tree-lstm}
    h^\smsuperscript{T}_j = \mathrm{TreeLSTMCell}(e^{(r)}_j, \{h^\smsuperscript{T}_k\}_{k\in C_j}) ,
\end{equation}
here $e^{(r)}_j$ denotes the encoding of relation triple $r_j$, $h^\smsuperscript{T}_j$ is the hidden states of $o_j$, and $C_j$ is the set of all child nodes of $o_j$.
Due to space limits, we refer readers to the supplementary material\nop{\citet{tai2015improved}} for more details about the Tree-LSTM architecture and the special handling of leaf and root nodes.

\noindent\textbf{Encode relations.}
Now we clarify how a relation triple $r$=$(s, p, o)$ is encoded. 
A set of features for each relation triple is created as shown in Table \ref{tab:relation-features}.
For $s$ and $o$, we use their entity type embeddings\nop{\footnote{Each Freebase predicate has a pair of designated type signatures for its connected entities \cite{berant2013semantic}.}}, entity type names\nop{\footnote{Retained by Freebase predicate: \tt{type.object.name}}} and their node types in the query graph.
For predicate $p$, we first find its inverse relation $\string^p$ and then use the embeddings and names for both of them.
The name features $n^{(s)}, n^{(p)}, n^{(\string^p)}, n^{(o)}$ are joined into a token sequence separated by a specialized delimiter. It is then encoded by an LSTM network. Next, the LSTM final hidden states are concatenated with the embedding vector features $t^{(s)}$, $v^{(s)}$, $t^{(p)}$, $t^{(\string^p)}$, $t^{(o)}$, $v^{(o)}$ and passed through a feed-forward neural network to get the final relation encoding $e^{(r)}$.

We do not use the vector representations of specific KG entities as in previous works for two reasons:
First, using the type information of entities as input should be enough because the task is to generate placeholders rather than specific entity names.
Second, our approach does not require the question and variable nodes in the query graph to be matched with entities that already exist in the KG or have trained embeddings as previous works do, allowing it to be more generally applicable\nop{ (e.g., it can generate questions for KG expansion)}.

\subsection{Decoder} 
At each decoding time step, the decoder generates either a word from the vocabulary, or a placeholders corresponding to an entity in the query graph.

Before introducing the placeholder generation strategy, we first clarify how to implement a standard LSTM decoder with attention \cite{bahdanau2014neural} over relation triples. A coverage mechanism \cite{see2017get} is also employed to encourage the decoder to generate tokens in a way such that all relation triples can be covered. The hidden states of the decoder LSTM are initialized by those of the root node from the Tree-LSTM encoder.
At time step $t$,
\begin{align}
    h_t &= \mathrm{LSTMCell}(x_{t-1}, h_{t-1}) ,\\
    c_{tj} &= \textstyle\sum_{t'=0}^{t-1}a_{t'j}, \\
    a_{tj} &\propto {w^{\scaleto{(1)}{6pt}}}^{T} \mathrm{tanh}(W^{\scaleto{(1)}{6pt}}[h_t, e^{(r)}_j, c_{tj}]+b^{\scaleto{(1)}{6pt}}) , \label{eq:attn_over_rel}\\
    r^*_t &= \textstyle\sum_{j} a_{tj} e^{(r)}_j ,
\end{align}
where $h_{t}$ is the previous hidden state of the decoder LSTM, $x_{t-1}$ is the word embedding of the previously generated token, $a_{tj}$ ($c_{tj}$) is the attention (coverage) on relation $r_j$ at time step $t$, and $r^*_t$ is the context vector. The probability distribution of words in the vocabulary to be generated is
\begin{equation}
\label{eq:p_vocab}
    P_\text{vocab} = \mathrm{softmax}(W^{\scaleto{(2)}{6pt}}[h_t, r^*_t] + b^{\scaleto{(2)}{6pt}}).
\end{equation}

\begin{table}[t!]
\centering
\caption{List of features representing a relation triple $r=(s,p,o)$. 
Node type $v^{(\cdot)}$$\in$$\{question, variable, terminal\}$ as discussed in Section~\ref{sec:preliminary}.
}
\resizebox{0.9\linewidth}{!}{
\begin{tabular}{@{}lll@{}}
                    & \textbf{Type}     & \textbf{Description}   \\ \midrule
$t^{(s)}$             & Embedding vector  & Entity type of subject node $s$.  \\
$n^{(s)}$             & String            & Canonical name of $t^{(s)}$. \\
$v^{(s)}$             & Embedding vector  & Node type of $s$ in the query graph. \\
\midrule
$t^{(p)}$             & Embedding vector  & Predicate $p$.    \\
$n^{(p)}$             & String            & Canonical name of $p$. \\
$t^{(\string^p)}$       & Embedding vector  & Inverse predicate $\string^p$.  \\
$n^{(\string^p)}$       & String            & Canonical name of $\string^p$.  \\
\midrule
$t^{(o)}$             & Embedding vector  & Entity type of object node $o$.   \\
$n^{(o)}$             & String            & Canonical name of $^{(s)}$. \\
$v^{(o)}$             & Embedding vector  & Node type of $o$ in the query graph. \\
\bottomrule
\end{tabular}
}
\label{tab:relation-features}
\end{table}
\noindent\textbf{A new placeholder generation strategy.}
Placeholders cannot be generated in the same way as vocabulary words because they represent different entities in different query graphs and do not have a global meaning. Observing that each relation triple contains at most one grounded entity, we design a new strategy that regulates the probability of generating a placeholder to be proportional to the attention over its related relation triples.
In addition, $p_\text{ph}\text{$\in$}[0,1]$ is the output of a sigmoid function, and it is used to control whether to generate a placeholder or a word from the vocabulary:
\begin{align}
    &p_\text{ph} = \sigma(W^{\scaleto{(3)}{6pt}}[x_t, h_t, r^*_t] + b^{\scaleto{(3)}{6pt}}) , \label{eq:p_ph}\\
    &P(w)\text{$=$}
    \begin{cases}
        (1\text{$-$}p_\text{ph}) P_\text{vocab}(w) & w\text{$\in$}\text{vocab} \\
        p_\text{ph} \alpha'_{tj} & w\text{$=$}\text{ph}(r_j) \label{eq:prob_placeholder}
    \end{cases}.
\end{align}
Note that $\alpha_{ij}'$ here differs from $\alpha_{ij}$ in Eqn.\ref{eq:attn_over_rel}, as it is the softmax over only those relations containing a grounded entity instead of all the relations.

\subsection{Training}
\label{sec:t2s_train}
\textsf{\small CoG2Q} can be trained with Maximum Likelihood Estimation to optimize the probability of generating the ground truth token $w^*$, i.e., $P(w^*)$ from Eqn.~\ref{eq:prob_placeholder}, at each time step.
We also add a coverage loss $\mathcal{L}_\text{COV}$ \cite{see2017get} to encourage attending over all the relations, and an L2 regularizer on the model parameters:
\begin{align}
    &\mathcal{L}_\text{MSE} = -\text{log}(P(w^*)), \label{eq:l_mse}\\
    &\mathcal{L}_\text{COV} = \textstyle\sum_{j} \text{min}(a_{tj}, c_{tj}) , \label{eq:l_cov}\\
    &\mathcal{L} = \mathcal{L}_\text{MSE} + \lambda_1 \mathcal{L}_\text{COV} + \lambda_2 L2 . \label{eq:l}
\end{align}

\section{\textsf{CoGSub2Q}: Sub-question Aided CQG}
\label{sec:incorp-subq}
A straightforward way to leverage sub-questions {(Def.\ref{def:subq})} of a given query graph is data augmentation, i.e., using them as additional training samples to train the \textsf{\small CoG2Q} model. 
However, this strategy does not fully utilize the intrinsic connections within the datasets: It can only rely on the trained parameters contributed by all the sub-questions (\emph{model-level} connection), rather than the specific sub-question corresponding to a given complex query graph (\emph{instance-level} connection).

Therefore, we propose to explicitly take sub-questions as extra inputs in the generation process, and extend the \textsf{\small CoG2Q} model with a sub-question encoder and a copying mechanism \cite{see2017get} in the decoder.
We refer to this model as \textsf{\small CoGSub2Q}. 
After encoding a given query graph and a sub-question, the new decoder of \textsf{\small CoGSub2Q} generates from an interpolated probability distribution of sub-question words, vocabulary words and placeholders.
More concretely, a sub-question token sequence $[z_1 ... z_N]$ is encoded by an LSTM encoder into vector representations $[h^\smsuperscript{Q}_1 ... h^\smsuperscript{Q}_N]$, and then an attention mechanism is applied over them, excluding $h^\smsuperscript{Q}_i$ if $z_i$ is a placeholder:
\begin{align}
    &[h^\smsuperscript{Q}_0...h^\smsuperscript{Q}_N] = \mathrm{LSTM}^\smsuperscript{Q}([z_0  ... z_{N}]) , \label{eq:enc-subq} \\
    &a^\smsuperscript{Q}_{ti} \propto w^T\text{tanh}(W^{\scaleto{(4)}{6pt}}[h_t, h^\smsuperscript{Q}_i] + b^{\scaleto{(4)}{6pt}}) , \label{eq:attn_over_subq}\\
    &{h^\smsuperscript{Q}_t}^* = \textstyle\sum_{j}a^\smsuperscript{Q}_{tj}h^\smsuperscript{Q}_j .
\end{align}

Intuitively, a complex question may reuse the expressions in a sub-question, so we allow the decoder to directly copy words from a sub-question.
We define $p_\text{copy}$ as the probability of copying a word from a sub-question rather than generating it from the entire vocabulary. 
Overall, the probability of outputting a vocabulary word is the sum of directly generating it from the vocabulary and copying it from the sub-question, if it appears (Eqn.\ref{eq:p_vocab_prim_prim}).
The other parts of the model are the same as \textsf{\small CoG2Q}, including the attention over relations and the generation of placeholders.
Overall, the generation probability $P(w)$ follows:
\begin{align}
    &p_\text{copy} = \sigma(W^{\scaleto{(5)}{6pt}}[x_t, h_t, {h^\smsuperscript{Q}_t}^*] + b^{\scaleto{(5)}{6pt}}) , \label{eq:p_copy} \\
    &P'_\text{vocab} = \text{softmax}(W^{\scaleto{(7)}{6pt}}[h_t, r^*_t, {h^\smsuperscript{Q}_t}^*] + b^{\scaleto{(7)}{6pt}}) \label{eq:p_vocab_prim} \\
    \begin{split}
    P_\text{vocab}^\text{sub}(w) = (1-p_\text{copy})P'_\text{vocab}(w) + \\ 
    p_\text{copy}\textstyle\sum_{j:z_j=w}a^\smsuperscript{Q}_{tj} ,
        \label{eq:p_vocab_prim_prim}
    \end{split} \\
        &p'_\text{ph} = \sigma(W^{\scaleto{(6)}{6pt}}[x_t, h_t, r^*_t, {h^\smsuperscript{Q}_t}^*] + b^{\scaleto{(6)}{6pt}}) , \label{eq:p_ph_prim} \\
    &P(w)\text{$=$}
    \begin{cases}
        (1\text{$-$}p'_\text{ph}) P_\text{vocab}^\text{sub}(w) & w\text{$\in$}\text{vocab} \\
        p'_\text{ph} \alpha'_{tj} & w\text{$=$}\text{ph}(r_j)
    \end{cases}. \label{eq:prob_placeholder_prim}
\end{align}

Based on the description of WebQ and CompQ in Section~\ref{sec:dataset}, there is a one-to-one correspondence between a complex question and a sub-question, as the annotators for CompQ refer to a sub-question while composing a complex question.
We assume this information is known during the training phase and train \textsf{\small CoGSub2Q} in the same way as described in Section~\ref{sec:t2s_train} (Eqn.\ref{eq:l_mse} -\ref{eq:l}).
At testing time, given a complex query graph in CompQ, we randomly sample one of its pseudo sub-questions (if there are many), use the trained \textsf{\small CoGSub2Q} model to generate a complex question and evaluate its quality with multiple ground-truth questions as references.
\section{\textsf{CoGSub$^m$2Q}: CQG with Multiple Pseudo Sub-Questions}
\label{sec:pseudo-subq-cpq}
In reality, the corresponding sub-questions (Def.\ref{def:subq}) may not be available for a given query graph. However, there already exist large-scale simple question datasets such as SimpQ discussed in Section~\ref{sec:dataset}, and moreover, it can be easier to harvest simple questions from user queries, chat logs, etc.
Such simple questions can be treated as pseudo sub-questions (Def.\ref{def:pseudo-subq}) for a complex query graph as long as the predicate in the former appears in the latter.
Therefore in this section, we discuss how to leverage pseudo sub-questions to help CQG, in the scenario where sub-questions are not available.

Since we can usually find multiple pseudo sub-questions for a given query graph, we further extend \textsf{\small CoGSub2Q} to automatically learn which pseudo sub-question is better associated with the query graph and can be more helpful for generating the ground-truth question. 
We refer to this model as \textsf{\small CoGSub$^m$2Q}.
Formally, given a set of $M$ simple questions $\{[z^m_1...z^m_{N^m}]\}_{m \in [1:M]}$, each of them is used independently as before in Section~\ref{sec:incorp-subq}, and we use $P_m(w)$ (corresponding to $P(w)$ in Eqn.\ref{eq:prob_placeholder_prim}) to represent the overall generation probability using the $m$-th pseudo sub-question.

We then define a two-level attention scorer module as follows:
First, $a_{jm}$ is the attention over the $j$-th word in the $m$-th simple question, computed from the simple question encoding $[h^{{\smsuperscript{Q}}_m}_1...{h^{{\smsuperscript{Q}}_m}_{N_m}}]$ (Eq.\ref{eq:enc-subq}) and the Tree-LSTM encoding of root node denoted as $h^\smsuperscript{T}_{root}$\nop{ (Eq.\ref{eq:tree-lstm})}.
Then, $a_m$ is the attention score of the $m$-th simple question as a whole, computed from $h^\smsuperscript{T}_{root}$ and the attentive aggregation of simple question word encodings. Formally,
\begin{align}
    &a_{jm} \propto W^{\scaleto{(8)}{6pt}} [h^{{\smsuperscript{Q}_m}}_j, h^\smsuperscript{T}_{root}] , \\
    &{h^{{\smsuperscript{Q}_m}}}^* = \textstyle\sum_j a_{jm} h^{{\smsuperscript{Q}_m}}_j , \\
    &a_{m} \propto W^{\scaleto{(9)}{6pt}} [{h^{{\smsuperscript{Q}_m}}}^*, h^{\smsuperscript{T}}_{root}] .
\end{align}
The predicted attention scores can be used to compute the weighted average of generating with different pseudo sub-questions. Specifically,
\begin{align}
    \bar{P}(w) = \textstyle\sum_m a_m P_m(w).
\end{align}

\section{Experiments}
\label{sec:experiment}
We conduct extensive experiments regarding the effectiveness of \textsf{\small CoGSub2Q} and \textsf{\small CoGSub$^m$2Q} using simple questions for CQG (Section~\ref{sec:exp-TS2S}). Before that, let us first show the competitiveness of our base model \textsf{\small CoG2Q} in Section~\ref{sec:exp-T2S}.

Following previous works, we employ BLEU\footnote{Due to space limits, only BLEU-4 is shown in the tables below but we observe similar results on BLEU 1-3, which are included in the supplementary material.} \cite{Papineni2001BleuAM}, METEOR \cite{Banerjee2005METEORAA} and ROUGE-L \cite{Lin2004ROUGEAP} as automatic evaluation metrics, and discuss human evaluation results in Section~\ref{sec:exp-userstudy}. 
{These automatic metrics are computed on questions with placeholders to avoid the score inflation from long entity names.}

\begin{table}[t!]
\centering
\caption{Compare different models and ablations that only use a KG query graph as input for CQG.\nop{without and with associated sub-questions from WebQ}}
\resizebox{0.9\linewidth}{!}{
\begin{tabular}{@{}p{3.3cm}ccc@{}}
                                            & \textbf{BLEU-4} & $\text{\textbf{ROUGE}}_\text{\textbf{L}}$   & \textbf{METEOR}    \\ \midrule
Enc-Dec                                 & 22.85 & 47.10                     & 32.78     \\
G2S \small{\cite{song-etal-2018-graph}} & 24.90 & 46.50                     & 35.54          \\
\textsf{CoG2Q}                   & \textbf{28.68} & \textbf{51.83}   & \textbf{37.54}     \\
\midrule
\textsf{CoG2Q} - attn            & 24.22 & 48.23                     & 35.56     \\
\textsf{CoG2Q} - $\string^p$      & 27.94 & 51.12                     & 36.84     \\
\textsf{CoG2Q} - names           & 28.11 & 51.36                     & 36.70     \\
\bottomrule
\end{tabular}
}
\label{tab:compare_encoder_less}
\end{table}
\subsection{Base Model Comparison}
\label{sec:exp-T2S}
We compare our base model \textsf{\small CoG2Q} with some of the state-of-the-art models to show its competitiveness:

(1) The \textbf{Enc-Dec} baseline is adapted from the previous simple question generation work \cite{serban2016generating}, where the query graph is simplified as a bag of relation features that are built from $(s,p,o)$ triples the same way as \textsf{\small CoG2Q}, and the same decoder with relation attention and placeholder generation method is used. 

(2) The graph-to-sequence (\textbf{G2S}) model \cite{song-etal-2018-graph} is one of the state-of-the-art models from the AMR-to-text generation literature \cite{Flanigan2016GenerationFA,Konstas2017NeuralAS}.
We also adapt it to take as input the same set of features, and to use the exact same decoder as \textsf{\small CoG2Q}\nop{, but with a graph encoder versus the tree-LSTM encoder}.

As shown in Table~\ref{tab:compare_encoder_less}, our base model \textsf{\small CoG2Q} outperforms all baselines, indicating the effectiveness of employing Tree-LSTM to aggregate the semantic information of a query graph from grounded entities to the question node.
It also indicates that sub-questions are not proximal to the target complex questions.

We also conduct an ablation study for \textsf{\small CoG2Q}\nop{, by removing the attention over relations, inverse predicate features, or the name features of the relations}, and the lower parts of Table~\ref{tab:compare_encoder_less} show the results.
We can see that the attention mechanism over relation triples is important, as it allows the decoder to directly access the relation encodings, \nop{I prefer we rephrase this: creates a highway path over Tree-LSTM directly from relation encodings to the decoder.}
and focus on expressing different relations at different time steps.
In addition, we observe that the overall performance improves by: (1) incorporating the feature vector of a predicate when it appears in the inverse direction, and (2) encoding the canonical names of the entity types and predicates.

\subsection{Leveraging Simple Questions}
\label{sec:exp-TS2S}
Given the strong base model \textsf{\small CoG2Q}, we now compare it with its extensions, in terms of how effectively they utilize simple questions.
Note that only pseudo sub-questions are assumed at test time for both \textsf{\small CoGSub2Q} and \textsf{\small CoGSub$^m$2Q}.
So for these two models, we run the testing five times and report the average and standard deviation results of the automatic metrics.

\noindent\textbf{Using sub-questions.}
The experiment results in Table~\ref{tab:compare_using_subq} show that sub-questions from WebQ can help complex question generation.
In particular, we make the following observations:
(1) The straightforward data augmentation strategy (\textbf{+ data}) improves the performance of \textsf{\small CoG2Q} by incorporating the sub-questions as extra training samples.
Note that for fair comparison, we have used all the sub-questions (corresponding to not only training, but also validation and testing complex questions) as extra training samples.
(2) The \textbf{SubQ} baseline simply compares the closeness of a sub-question to a complex question. This shows that \textsf{\small CoGSub2G} needs to do more than just copy every word from the sub-questions. 
(3) More importantly, we hypothesize in Section~\ref{sec:incorp-subq} that it will be more effective to leverage the connections between sub-questions and complex questions on the \emph{instance-level} than the \emph{model-level}.
It is verified by the results that data augmentation is outperformed by \textsf{\small CoGSub2Q}, which can explicitly encode and copy from sub-questions.
(4) Very interestingly, we also discover that the sub-question correspondence information is essential to train the \textsf{\small CoGSub2Q} model successfully.
Specifically, we experiment with \textsf{\small CoGSub2Q} in two different scenarios, one trained with randomly sampled pseudo sub-questions and the other with actual sub-questions.
We can see from Table~\ref{tab:compare_using_subq} that \textsf{\small CoGSub2Q} performs similarly to data augmentation when trained with pseudo sub-questions, but significantly better with actual sub-questions.
The reason for this is similar to why data augmentation is not optimal:
the sub-question to complex question mapping is less clearly established when the model learns from pseudo sub-questions instead of actual ones.
Conversely, learning with sub-questions better exploits the instance-level connections, and it is easier for the model to learn with the actual sub-questions that often contain more words appearing in the target complex questions.
\begin{table}[t!]
\centering
\caption{
CQG with sub-questions from WebQ.
Superscript ``pseudo" indicates training with randomly sampled pseudo sub-questions rather than real ones.
* denotes significantly different from \scaleto{\textsf{\small CoG2Q}}{6pt}+data in one-tailed t-test (p$<$0.05).}
\resizebox{0.98\linewidth}{!}{
\begin{tabular}{@{}p{2.7cm}lll@{}}
& \textbf{BLEU-4} & $\text{\textbf{ROUGE}}_\text{\textbf{L}}$  & \textbf{METEOR}    \\ 
\midrule
SubQ                    & 15.89 & 33.87                     & 19.62     \\    
\textsf{CoG2Q}          & 28.68 & 51.83                     & 37.54     \\\midrule
\textsf{CoG2Q} + data   & 29.23 & 52.64                     & 38.33     \\ 
\textsf{CoGSub2Q}$^\text{pseudo}$       & 29.23$\pm$0.14 & 52.68$\pm$0.12 & 37.98$\pm$0.12     \\  
\textsf{CoGSub2Q}       & \textbf{31.43}$\pm$0.13$^*$ & \textbf{55.26}$\pm$0.13$^*$ & \textbf{39.77}$\pm$0.06$^*$     \\
\bottomrule
\end{tabular}
}
\label{tab:compare_using_subq}
\end{table}

\noindent\textbf{Using multiple pseudo sub-questions.}
In the scenario when the sub-questions are not available, we now demonstrate whether pseudo sub-questions from SimpQ can be used as surrogates of real sub-questions to help complex question generation.
Table~\ref{tab:compare_using_pseudo} shows the results
\footnote{We only keep the complex questions with a predicate appearing in SimpQ for fair comparison.}.
\begin{table}[t!]
\centering
\caption{
    CQG with multiple pseudo sub-questions from SimpQ. 
    (See Table~\ref{tab:compare_using_subq} for meaning of ``*".)
}
\resizebox{0.98\linewidth}{!}{
\begin{tabular}{@{}p{2.7cm}lll@{}}
                            & \textbf{BLEU-4} & $\text{\textbf{ROUGE}}_\text{\textbf{L}}$   & \textbf{METEOR}    \\ 
\midrule
\textsf{CoG2Q}          & 27.89 & 51.62                     & 36.50     \\ 
\midrule
\textsf{CoG2Q} + data   & 28.87 & 52.35                     & 37.83     \\
\textsf{CoGSub2Q}$^\text{pseudo}$      & 29.58$\pm$0.13 & 52.94$\pm$0.14 & 38.02$\pm$0.05 \\  
\midrule
\textsf{CoGSub$^m$2Q}   & \textbf{30.95}$\pm$0.19$^*$ & \textbf{53.68}$\pm$0.10$^*$ & \textbf{38.74}$\pm$0.05$^*$ \\
\bottomrule
\end{tabular}
}
\label{tab:compare_using_pseudo}
\end{table}

Using simple questions for data augmentation consistently improves the performance of the base model \textsf{\small CoG2Q}.
\nop{We test \textsf{\small CoGSub2Q} with one randomly sampled pseudo sub-question because real sub-questions are rare in SimpQ.}
The results are similar as before in that it does not perform significantly better than data augmentation.
\nop{I suggest we remove:\st{Its slight performance gain over data augmentation may be because questions from SimpQ are simpler than those in WebQ, therefore are easier to copy from but less useful for handling complex query graphs.}, and if we have some space, provide more analysis in user/case study about the generated questions (maybe refer to our ACL'19), rather than the scores here.}
Finally, our proposed \textsf{\small CoGSub$^m$2Q} model taking as input multiple simple questions shows significant performance gain over \textsf{\small CoG2Q} with data augmentation, proving that this model can simultaneously learn to predict higher association scores for more useful pseudo sub-questions and use these simple question for CQG.

\subsection{Human Evaluation and Case Study}
\label{sec:exp-userstudy}
{Although automatic evaluation metrics have been the most widely adopted and trusted in the QG literature \cite{serban2016generating,elsahar2018zero}, they still have some limitations.}
So as to complement the automatic evaluations, we conduct a user study to compare the data augmentation method with our models that explicitly leverage (pseudo) sub-questions.
Four human annotators \nop{who are all proficient English speakers }are asked to judge a generated question after seeing the corresponding query-graph and ground-truth complex questions.
We randomly sample 100 examples for each scenario with sub-questions or pseudo sub-questions.
The generated complex questions are judged based on \textit{correctness} and \textit{naturalness}.
{We instruct the annotators to count each relation or two relations connected with a CVT node as a single constraint.}
Correctness is defined as\nop{The former is} the fraction of constraints\nop{\footnote{We instruct the annotators to count each relation or two relations connected with a CVT node as a single constraint.}} that have been expressed correctly, and naturalness\nop{the latter} is a score ranging between 1 and 5.
Table~\ref{tab:user_study} shows the results.
In both scenarios, our methods outperform the data augmentation counterpart in terms of correctness, indicating that they can more effectively leverage the (pseudo) sub-questions.
However, they do not achieve much higher naturalness scores.
This is probably because, although directly copying words from human written (pseudo) sub-questions may contribute to higher naturalness, the fusion of copied words and generated words is prone to unnatural connections.

We perform an error analysis on the human scored questions and identify four major error types as shown in Table~\ref{tab:error_analysis}.
Some questions have more than one types of error.
The most dominant error type is missing a relation.
This suggests that we need a better method than the exiting coverage mechanism to encourage describing all relations.
There are also many cases where a relation is expressed with irrelevant words, or the placeholder of a wrong entity is generated. Such errors may stem from rare relation combinations, or imperfect attention over relations.
A smaller amount of errors are repeated relations, which may also be caused by the above reasons.

Due to space limit, we only show a case study of utilizing pseudo sub-questions in Table~\ref{tab:case_study} and put more examples in the supplementary material.
Here, question node $x$ is connected through inverse \texttt{\small capital} predicate to $c$, which is further connected to CVT node $k$ and then to \texttt{\small apple} growing industry (not the company).
The generation from \textsf{\small CoG2Q} neglects the ``capital" relation and falsely described the ``export" relation, probably because its decoder gets confused by the combination of these two constraints that never appear during training.
On the contrary, \textsf{\small CoGSub$^m$2S} successfully ranks the most relevant pseudo sub-question ``\textit{what is the capital of \textlangle{}ent\textrangle{}}" at the top, and is able to borrow words from it except the placeholder. 
It can then focuses on describing the ``export" relationship and generate the ``\textit{country whose exports is \textlangle{}ent\textrangle{}}" clause.
\begin{table}[t]
\centering
\caption{Human evaluation results in both scenarios.}
\resizebox{0.95\linewidth}{!}{
\begin{tabular}{@{}p{1.6cm}lcc@{}}
&                                           & \textbf{Correctness}   & \textbf{Naturalness}  \\ 
\midrule
\multirow{2}{*}{\renewcommand{\arraystretch}{0.7}\begin{tabular}[c]{@{}c@{}}\small{\textbf{w/ sub-}}\\\small{\textbf{questions}}\end{tabular}}
&\textsf{CoG2Q} + data      & 0.5242        & 3.24     \\    
&\textsf{CoGSub2Q}          & 0.6218        & 3.28     \\ 
\midrule
\multirow{2}{*}{\renewcommand{\arraystretch}{0.7}\begin{tabular}[c]{@{}c@{}}\small{\textbf{w/ pseudo}}\\\small{\textbf{sub-questions}}\end{tabular}}
&\textsf{CoG2Q} + data      & 0.4817        & 3.42     \\    
&\textsf{CoGSub$^{m}$2Q}    & 0.5717        & 3.52     \\
\bottomrule
\end{tabular}
}
\label{tab:user_study}
\end{table}
\begin{table}[t]
\centering
\caption{Percentage of different types of errors among all.}
\resizebox{0.5\linewidth}{!}{
\begin{tabular}{@{}lc@{}}
\textbf{Error Type} & \textbf{Percentage}   \\ 
\midrule
Missing relation    & 53\%  \\ 
Repeated relation   & 15\%  \\
Wrong relation      & 32\%  \\
Wrong entity        & 25\%  \\
\bottomrule
\end{tabular}
}
\label{tab:error_analysis}
\end{table}

\begin{table}[t]
\centering
\caption{Case study. Pseudo sub-questions help CQG.}
\resizebox{0.99\linewidth}{!}{
    \begin{tabular}{@{}ll@{}}
    \multicolumn{2}{c}{
        \includegraphics[width=90mm]{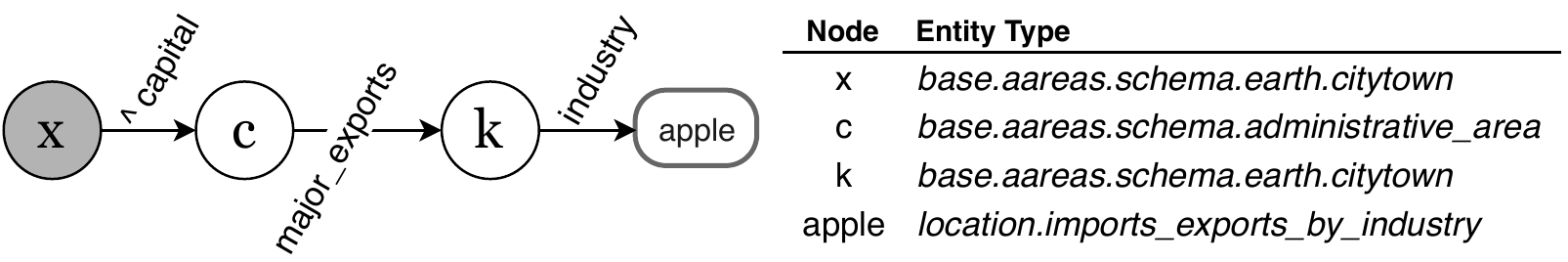}
        }  \\[-1mm]
    \midrule
    Ground-truth & what is the state capital where the major export is \textbf{apple} trees ?  \\
    \midrule
    \textsf{CoG2Q} + data & what is the name of the country where the \textbf{apple} is located?  \\
    \midrule
    \textsf{CoGSub$^m$2Q} & what is the capital of the country whose exports is \textbf{apple}?  \\
    \cmidrule{2-2}
    \makecell[cl]{Ranked pseudo\\sub-questions} & \makecell[cl]{what is the capital of the \textlangle{}ent\textrangle{} ? \\ what is the capital of the administrative area of \textlangle{}ent\textrangle{}? \\ what is the capital of \textlangle{}ent\textrangle{}? \\ what's the capital of \textlangle{}ent\textrangle{}? \\ what is the capital of \textlangle unk\textrangle? ... } \\
    \bottomrule
    \end{tabular}
}
\label{tab:case_study}
\end{table}
\section{Related Work}
\noindent\textbf{Question generation.}
\citet{serban2016generating} first study QG from KG and design an encoder-decoder model to generate a question from a single relation triple. 
\citet{elsahar2018zero} further study simple question generation in zero-shot settings to handle unseen predicates or entity types by incorporating external context information. 
Some other works study (semi) automatic question generation from KG.
\citet{Song2016QuestionGF} propose a retrieval-based system to search the web for simple questions with seed questions generated by templates.
\citet{talmor2018web,Su2016OnGC} automatically find interesting query graphs, and then hire humans to translate them into natural language questions.
Moreover, question generation has been studied with data sources other than KG and for different purposes.
For example, questions have been generated from text paragraphs \cite{Du2017LearningTA,Song2018LeveragingCI}, images \cite{Mostafazadeh2016GeneratingNQ,Patro2018MultimodalDN}, or combined with question answering tasks \cite{Duan2017QuestionGF,Tang2017QuestionAA}.
To the best of our knowledge, we are the first to study complex question generation from KG, with a focus on leveraging existing simple questions.

\noindent\textbf{Natural language generation from structured inputs.}
Many works focus on generating natural language utterances from structured inputs, such as Resource Description Framework triples \cite{Trisedya2018GTRLSTMAT}, SQL \cite{Xu2018SQLtoTextGW}, or Abstract Meaning Representations, either with traditional methods \cite{Liu2015TowardAS,Flanigan2016GenerationFA,Song2016AMRtotextGA} or neural encoder-decoders \cite{Konstas2017NeuralAS,song-etal-2018-graph}.
Directly applying these similar but different techniques may not be ideal because the encoder for CQG needs to aggregate information towards the question nodes.
Also, KG makes it easy to find corresponding simple questions since the entities and predicates belong to a global ontology, which is absent in other structured inputs.

\section{Conclusion}
We study complex question generation from KG, with an emphasis on how to leverage existing simple questions.
A strong base neural encoder-decoder model that converts a query graph to a natural language question is first designed, and then we propose two extensions that explicitly consider instance-level connections between simple and complex questions, which are empirically shown to be more effective than straightforward data augmentation.

\bibliographystyle{aaai}
\bibliography{bibliography}

\section{Supplementary Material}
\label{sec:supplemental}
\begin{table*}[t!]
\centering
\small
\resizebox{0.6\linewidth}{!}{
\begin{tabular}{@{}p{3.3cm}cccccc@{}}
\toprule
                & \textbf{BLEU-1} & \textbf{BLEU-2} & \textbf{BLEU-3} & \textbf{BLEU-4} & $\text{\textbf{ROUGE}}_\text{\textbf{L}}$   & \textbf{METEOR}    \\ \midrule
Enc-Dec         & 54.50 & 38.29 & 28.46 & 22.85 & 47.10                     & 32.78     \\
G2S \cite{song-etal-2018-graph}      & 53.41 & 37.79 & 29.58 & 24.90 & 46.50 & 35.54 \\
\textsf{\small Cog2Q}             & 59.86 & \textbf{44.14} & \textbf{34.60} & \textbf{28.68} & \textbf{51.83}                     & \textbf{37.54}     \\
\midrule
\textsf{\small Cog2Q} - attn      & 56.70 & 40.18 & 29.99 & 24.22 & 48.23                     & 35.56     \\
\textsf{\small Cog2Q} - $\hat{} p$      & 59.53 & 43.43 & 33.66 & 27.94 & 51.12                     & 36.84     \\
\textsf{\small Cog2Q} - names         & \textbf{60.06} & 43.97 & 34.09 & 28.11 & 51.36                     & 36.70     \\
\bottomrule
\end{tabular}
}
\caption{Compare different models and ablations that onlyuse a KG query graph as input for CQG.}
\label{tab:compare_encoder_all}
\end{table*}
\begin{table*}[t]
\centering
\resizebox{0.65\textwidth}{!}{
    \begin{tabular}{@{}p{0.2cm}p{3cm}llllll@{}}
    \toprule
    & & \textbf{BLEU-1} & \textbf{BLEU-2} & \textbf{BLEU-3} & \textbf{BLEU-4} & $\text{\textbf{ROUGE}}_\text{\textbf{L}}$   & \textbf{METEOR} \\ 
    \toprule
    \parbox[t]{2mm}{\multirow{5}{*}{\rotatebox[origin=c]{90}{{\textbf{w/ sub-q}}}}}
    & SubQ                  & 32.10 & 22.11 & 17.79 & 15.89 & 33.87                     & 19.62     \\    
    &\textsf{CoG2Q}         & 59.86 & 44.14 & 34.60 & 28.68 & 51.83                     & 37.54     \\\cmidrule{2-8}
    &\textsf{CoG2Q} + data  & 60.72 & 45.03 & 35.38 & 29.23 & 52.64                     & 38.33     \\ 
    &\textsf{CoGSub2Q}$^\text{pseudo}$    & 61.33$\pm$0.16 & 45.11$\pm$0.09 & 35.32$\pm$0.14 & 29.23$\pm$0.14 & 52.68$\pm$0.12 & 37.98$\pm$0.12     \\  
    &\textsf{CoGSub2Q}       & \textbf{63.55}$\pm$0.04 & \textbf{47.83}$\pm$0.10 & \textbf{37.80}$\pm$0.13 & \textbf{31.43}$\pm$0.13 & \textbf{55.26}$\pm$0.13 & \textbf{39.77}$\pm$0.06     \\ \midrule\midrule
    
    \parbox[t]{2mm}{\multirow{5}{*}{\rotatebox[origin=c]{90}{{\textbf{w/ pseudo}}}}}
    &\textsf{CoG2Q}          & 59.46 & 43.96 & 33.89 & 27.89 & 51.62                     & 36.50     \\ \cmidrule{2-8}
    &\textsf{CoG2Q} + data   & 60.75 & 44.85 & 34.82 & 28.87 & 52.35                     & 37.83     \\
    &\textsf{CoGSub2Q}$^\text{pseudo}$ & 61.26$\pm$0.14 & 45.59$\pm$0.19 & 35.77$\pm$0.17 & 29.58$\pm$0.13 & 52.94$\pm$0.14 & 38.02$\pm$0.05 \\  
    \cmidrule{2-8}
    &\textsf{CoGSub$^m$2Q}   & \textbf{61.53}$\pm$0.11 & \textbf{46.47}$\pm$0.17 & \textbf{36.93}$\pm$0.18 & \textbf{30.95}$\pm$0.19 & \textbf{53.68}$\pm$0.10 & \textbf{38.74}$\pm$0.05\\
    \bottomrule
    \end{tabular}
}
\caption{CQG with sub-questions from WebQ (upper half) and multiple pseudo sub-questions from SimpQ (lower half).
Superscript ``pseudo" indicates training with randomly sampled pseudo sub-questions rather than real ones.}
\label{tab:compare_using_multiple_subquestions_all}
\end{table*}

\subsection{Implementation Details}
We use PyTorch\footnote{\url{https://pytorch.org/docs/0.4.1/}} to implement our models, and the hyper-parameters are selected based the on the dev set.
The hidden size of the Tree-LSTM encoder is set to 500.
The decoder LSTM and the (pseudo) sub-question encoder are both standard LSTM with a single layer, whose hidden sizes are set to 500 and 300 respectively.
The embedding vector sizes of words, entity types, entity node types and predicates are all 300. 
The relation encoding vector size is 500.
We train our model with the Adam optimizer with learning rate set at 1e-4 and decay rate at 0.8.
The weights on the coverage loss and the L2 parameter regularizer are 0.1 and 1e-4 respectively.

The entity types for query graph nodes are determined by the predicates. In Freebase, each predicate has a pair of designated type signatures for its connected entities \cite{berant2013semantic}.
The entity type names are retained by a special Freebase predicate \texttt{type.object.name}.

In the experiments not using simple questions or using sub-questions, the train/dev/test splits contains 3962/1144/1069 samples from CompQ, corresponding to 396/90/86 samples from WebQ. 
These numbers slightly differ from those in Table~\ref{tab:data_stats} because some query graphs in WebQ are grouped with others in CompQ into a complex sample.
In the experiments using pseudo sub-questions, the train/dev/test split sizes are 3057/909/827, as we only consider those complex query graphs having predicates that appears in SimpQ.
Under each scenario, we use the trained \textsf{\small CoG2Q} parameters to initialize the extension models \textsf{\small CoGSub2Q} and \textsf{\small CoGSub$^m$2Q}.

\noindent\textbf{Tree-LSTM.} We have discussed in Section~\ref{T2S} that \textsf{\small CoG2Q} employs a Tree-LSTM network to encode a query query and $h^\smsuperscript{T}_j$ denotes the encoding of node $j$.
We use the child-sum Tree-LSTM version because the child nodes are not ordered and the number of children is unfixed:
\begin{equation*}
    \label{eq:tree-lstm-supplementary}
    h^\smsuperscript{T}_j = \mathrm{TreeLSTMCell}(e^{(r)}_j, \{h^\smsuperscript{T}_k\}_{k\in C_j}) .
\end{equation*}
The detailed process within this TreeLSTMCell is:
\begin{align*}
    h^{\sim}_j &= \textstyle\sum_{k\in C(j)}h^\smsuperscript{T}_k ,\\
    i_j &= \sigma(W^{(i)}[e^{(r)}_j, h^{\sim}_j]+b^{(i)}) ,\\
    f_{jk} &= \sigma(W^{(f)}e^{(r)}_j, h^{\sim}_j]+b^{(f)}) ,\\
    o_j &= \sigma(W^{(o)}[e^{(r)}_j, h_k]+b^{(o)}) ,\\
    u_j &= tanh(W^{(u)}[e^{(r)}_j, h^{\sim}_j]+b^{(u)}) ,\\
    c_j &= i_j \odot u_j + \textstyle\sum_{k\in C(j)} f_{jk} \odot c_k ,\\
    h^\smsuperscript{T}_j &= o-j \odot tanh(c_j) .
\end{align*}

Note that the terminal nodes and the root (question) nodes are two special cases.
Zero vectors are used as the hidden states of the nonexistent ``children'' of terminal nodes; and a dummy relation encoding vector $e^{(r)}_\text{dummy}$ is used for the root nodes.

\subsection{Quantitative Results}
Table~\ref{tab:compare_encoder_all} compares models that map complex query graphs to complex questions without accessing the simple questions.
Our proposed \textsf{\small CoG2Q} achieves the overall best performance, and is therefore used as base model in the following studies.
Table~\ref{tab:compare_using_multiple_subquestions_all} summarizes the comparison of the base and extension models in both scenarios, utilizing sub-questions from WebQ or utilizing pseudo sub-questions from SimpQ.
As discussed earlier, the extension models outperform the data augmented base model under their respective scenarios, because they can utilize instance-level connections between simple and complex questions.

\begin{table*}[t]
\small
\centering
\resizebox{0.52\textwidth}{!}{
\begin{tabular}{@{}ll@{}}
\toprule
\multicolumn{2}{c}{\includegraphics[width=0.65\linewidth]{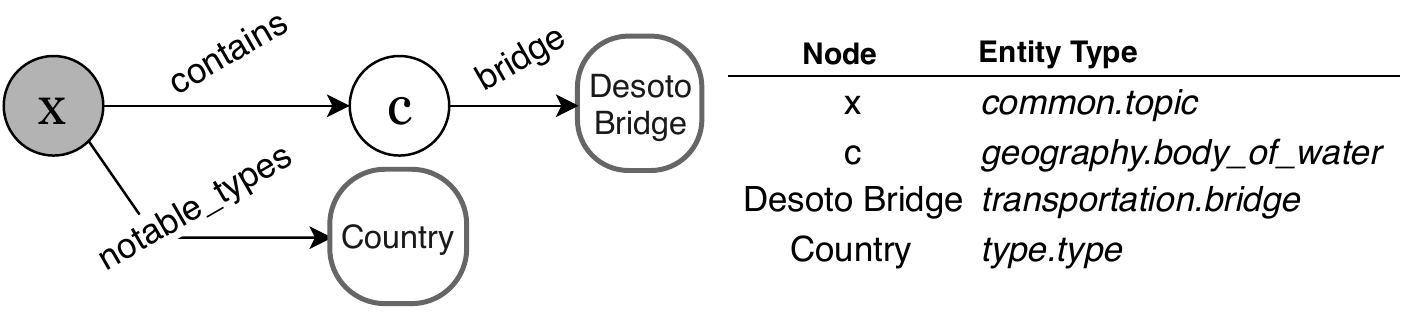}} \\
Ground-truth                             & what country is home to a body of water under \textbf{desoto bridge} ?  \\
\midrule
\textsf{\small CoG2S} + data         & what country is the river of the water of \textbf{desoto bridge} ?  \\
\midrule
\textsf{\small CoGSub2S}                 & what countries does the river of the \textbf{desoto bridge} run through ?  \\
\midrule
\makecell[cl]{Sub-questions}             & \makecell[cl]{what county is \textlangle{}ent\textrangle{} in ? \\ what country is \textlangle{}ent\textrangle{} located ? \\ what countries does the \textlangle{}ent\textrangle{} run through ? \\ what countries does the \textlangle{}ent\textrangle{} go through ? ... }\\
\midrule\midrule
\multicolumn{2}{c}{\includegraphics[width=0.9\linewidth]{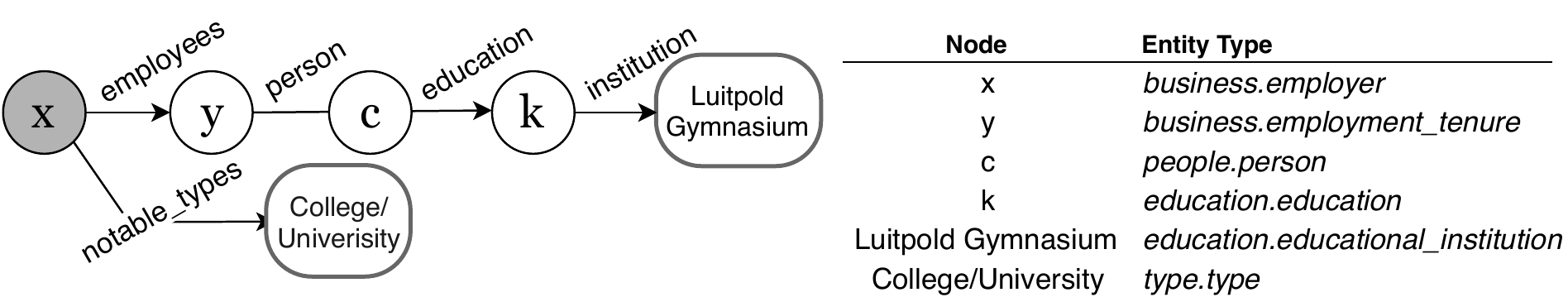}} \\
Ground-truth                             & the person with education institution \textbf{luitpold gymnasium} taught at what colleges? \\
\midrule
\textsf{\small CoG2S} + data         & what is the name of the person who attended the luitpold gymnasium who attended \textbf{luitpold gymnasium} ?
 \\
\midrule
\textsf{\small CoGSub2S}                 & what colleges did the person who attended \textbf{luitpold gymnasium} teach at ? \\
\midrule
\makecell[cl]{Sub-questions}             & \makecell[cl]{what colleges did \textlangle{}ent\textrangle{} teach at ?}\\
\bottomrule
\end{tabular}
}
\caption{Case study. \textsf{\small CoGSub2Q} using sub-questions.}
\label{tab:qualitative-subq}
\end{table*}

\begin{table*}[t]
\small
\centering
\resizebox{0.52\textwidth}{!}{
\begin{tabular}{@{}ll@{}}
\toprule
\multicolumn{2}{c}{\includegraphics[width=0.65\linewidth]{figures/case-study-w-type.pdf}} \\
Ground-truth                             & what is the state capital where the major export is \textbf{apple} trees ?  \\
\midrule
\textsf{\small CoG2S} + data         & what is the name of the country where the \textbf{apple} is located?  \\
\midrule
\textsf{\small CoGSub2S}                 & what is the capital of the country whose exports is \textbf{apple}?  \\
\midrule
\makecell[cl]{Ranked pseudo\\sub-questions} & \makecell[cl]{what is the capital of the \textlangle{}ent\textrangle{} ? \\ what is the capital of the administrative area of \textlangle{}ent\textrangle{}? \\ what is the capital of \textlangle{}ent\textrangle{}? \\ what's the capital of \textlangle{}ent\textrangle{}? \\ what is the capital of \textlangle{}unk\textrangle{}? ... }\\
\midrule\midrule
\multicolumn{2}{c}{\includegraphics[width=0.73\linewidth]{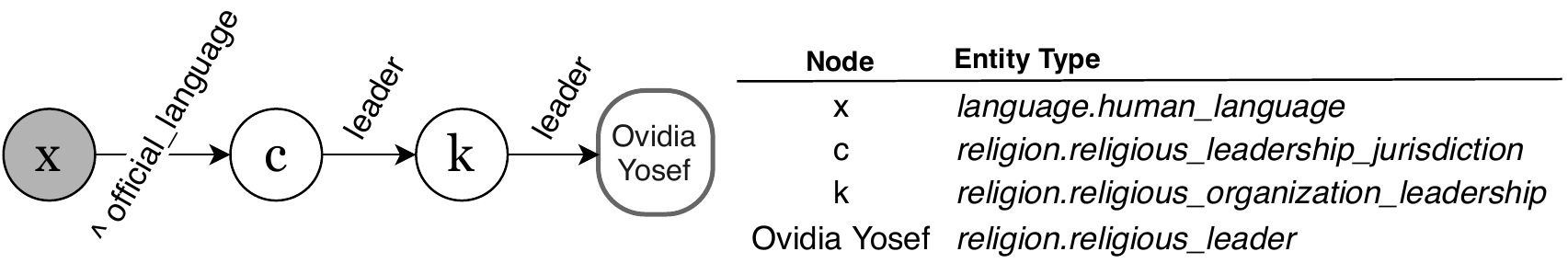}} \\
Ground-truth                             & what is the main language of the place with religious leader \textbf{ovadia yosef} ?  \\
\midrule
\textsf{\small CoG2S} + data         & what are the official language of the religious where \textbf{ovadia yosef} is a religious leader ? \\
\midrule
\textsf{\small CoGSub$^m$2S}             & what is the official language of the country that includes \textbf{ovadia yosef} as a religious leader ? \\
\midrule
\makecell[cl]{Ranked pseudo\\sub-questions}  & \makecell[cl]{what is the official language spoken in \textlangle{}ent\textrangle{}? \\ what is the official language of \textlangle{}ent\textrangle{}? \\ what 's the official language of \textlangle{}ent\textrangle{}? \\ what language do people speak in \textlangle{}ent\textrangle{}? \\ which language is spoken in \textlangle{}ent\textrangle{}? ...}\\
\bottomrule
\end{tabular}
}
\caption{Case study. \textsf{\small CoGSub$^m$2Q} using pseudo sub-questions.}
\label{tab:qualitative-pseudo}
\end{table*}
\subsection{More Case Studies}
Table~\ref{tab:qualitative-subq} and \ref{tab:qualitative-pseudo} show several testing examples of different query graphs.
We specifically compare the generations from the extension models with those from the data augmented base model, in order to provide some insights into how simple questions help complex question generation.
For instance, in the first example in Table~\ref{tab:qualitative-subq}, \textsf{\small CoGSub2Q} is able to borrow the phrase ``\textit{run through}" from a sub-question, which makes its generation both correct and specific to the fact that $c$ is a river.
The base model generation is also understandable, but it is less natural and too general, as it uses the word ``\textit{is}" to describe the ``contains" relationship between a country and a river.
In the second example, \textsf{\small CoGSub2Q} copies ``\textit{what college did ... teach at}" from the sub-question and describes the ``attend" relationship in between, whereas \textsf{\small CoG2Q} ignores the ``teach" relationship and describes the ``attend" relationship twice.

Table~\ref{tab:qualitative-pseudo} shows how \textsf{\small CoGSub2Q} leverages pseudo sub-questions.
The first example has been discussed in Section~\ref{sec:experiment} previously.
In the second example, the model copies ``\textit{what is the official language of}" from the pseudo sub-questions, and then focuses on the ``religious leader" relationship around CVT node $k$. In comparison, the base model successfully describes the ``official language" relationship as well but then gets confused and repeatedly generates ``\textit{of the religious}" and ``\textit{is a religious leader}".
\balance

\end{document}